\documentclass{article}

\usepackage{microtype}
\usepackage{graphicx}

\usepackage{float}
\usepackage{subcaption}

\usepackage{xfrac}
\usepackage{booktabs} 
\usepackage{amsmath}
\usepackage{amsfonts}
\usepackage{amssymb}

\usepackage{scalerel}

\usepackage{hyperref}

\usepackage[accepted]{icml2018}

\icmltitlerunning{Fast Information-theoretic Bayesian Optimisation}

\begin{document}

\twocolumn[
\icmltitle{Fast Information-theoretic Bayesian Optimisation}



\icmlsetsymbol{equal}{*}

\begin{icmlauthorlist}

\icmlauthor{Binxin Ru}{Ox}
\icmlauthor{Mark McLeod}{Ox}
\icmlauthor{Diego Granziol}{Ox}
\icmlauthor{Michael A. Osborne}{Ox,mf}
\end{icmlauthorlist}

\icmlaffiliation{Ox}{Department of Engineering Science, University of Oxford, Oxford, UK}
\icmlaffiliation{mf}{Mind Foundry Ltd., Oxford}

\icmlcorrespondingauthor{Binxin Ru}{robin@robots.ox.ac.uk}
\icmlcorrespondingauthor{Mark McLeod}{mark.mcleod@magd.ox.ac.uk}
\icmlcorrespondingauthor{Diego Granziol}{diego@robots.ox.ac.uk}
\icmlcorrespondingauthor{Michael A. Osborne}{mosb@robots.ox.ac.uk}

\icmlkeywords{Information theory, Bayesian Optimisation, Machine Learning, ICML}

\vskip 0.3in
]



\printAffiliationsAndNotice{}  

\begin{abstract}

Information-theoretic Bayesian optimisation techniques have demonstrated state-of-the-art performance in tackling important global optimisation problems. However, current information-theoretic approaches require many approximations in implementation, introduce often-prohibitive computational overhead and limit the choice of kernels available to model the objective. We develop a fast information-theoretic Bayesian Optimisation method, FITBO, that avoids the need for sampling the global minimiser, thus significantly reducing computational overhead. Moreover, in comparison with existing approaches, our method faces fewer constraints on kernel choice and enjoys the merits of dealing with the output space. We demonstrate empirically that FITBO inherits the performance associated with information-theoretic Bayesian optimisation, while being even faster than simpler Bayesian optimisation approaches, such as Expected Improvement.

\end{abstract}
\section{Introduction}
\label{introduction}

Optimisation problems arise in numerous fields ranging from science and engineering to economics and management \citep{brochu2010tutorial}. In classical optimisation tasks, the objective function is usually known and cheap to evaluate \citep{hennig2012entropy}. However, in many situations, we face another type of tasks for which the above assumptions do not apply. For example, in the cases of clinical trials, financial investments or constructing a sensor network, it is very costly to draw a sample from the latent function underlying the real-world processes \citep{brochu2010tutorial}. The objective functions in such type of problems are generally non-convex and their closed-form expressions and derivatives are unknown \citep{shahriari2016taking}. Bayesian optimisation is a powerful tool to tackle such optimisation challenges \citep{brochu2010tutorial}. 
\\\\
A core step in Bayesian optimisation is to define an acquisition function which uses the available observations effectively to recommend the next query location \citep{shahriari2016taking}. There are many types of acquisition functions such as Probability of Improvement (PI) \citep{kushner1964new}, Expected Improvement (EI) \citep{{movckus1978toward}, {jones1998efficient}} and Gaussian Process Upper Confidence Bound (GP-UCB) \citep{srinivas2009gaussian}. The most recent type is based on information theory and offers a new perspective to efficiently select the sequence of sampling locations based on entropy of the distribution over the unknown minimiser $\mathbf{x}_*$ \citep{shahriari2016taking}. The information-theoretic approaches guide our evaluations to locations where we can maximise our learning about the unknown minimum rather than to locations where we expect to obtain lower function values \citep{hennig2012entropy}. Such methods have demonstrated impressive empirical performance and tend to outperform traditional methods in tasks with highly multimodal and noisy latent functions.
\\\\
One popular information-based acquisition function is Predictive Entropy Search (PES) \citep{{villemonteix_informational_2009}, {hennig2012entropy}, {hernandez2014predictive}} . However, it is very slow to evaluate in comparison with traditional methods like EI, PI and GP-UCB and faces serious constraints in its application. For example, the implementation of PES requires the first and second partial derivatives as well as the spectral density of the Gaussian process kernel function \citep{{hernandez2014predictive},{requeimaintegrated}}. This limits our kernel choices. Moreover, PES deals with the input space, thus less efficient in higher dimensional problems \citep{wang2017max}. The more recent methods such as Output-space Predictive Entropy Search (OPES) \citep{hoffman:2015} and Max-value Entropy Search (MES) \citep{wang2017max} improve on PES by focusing on the information content in output space instead of input space. However, current entropy search methods, whether dealing with the minimiser or the minimum value, all involve two separate sampling processes : 1) sampling hyperparameters for marginalisation and 2) sampling the global minimum/minimiser for entropy computation. The second sampling process not only contributes significantly to the computational burden of these information-based acquisition functions but also requires the construction of a good approximation for the objective function based on Bochner's theorem \citep{hernandez2014predictive}, which limits the kernel choices to the stationary ones \citep{bochner1959lectures}. 
 \\\\ 
In view of the limitations of the existing methods, we propose a fast information-theoretic Bayesian optimisation technique (FITBO). Inspired by the Bayesian integration work in \citep{gunter2014sampling}, the creative contribution of our technique is to approximate any black-box function in a parabolic form: $ f(\mathbf{x}) =\eta+\sfrac{1}{2}g(\mathbf{x})^2 $. The global minimum is explicitly represented by a hyperparameter $\eta$, which can be sampled together with other hyperparameters. As a result, our approach has the following three major advantages: 
\begin{enumerate}
	\item Our approach reduces the expensive process of sampling the global minimum/minimiser to the much more efficient process of sampling one additional hyperparameter, thus overcoming the speed bottleneck of information-theoretic approaches. 	\item Our approach faces fewer constraints on the choice of appropriate kernel functions for the Gaussian process prior. 		\item Similar to MES \citep{wang2017max}, our approach works on information in the output space and thus is more efficient in high dimensional problems.
\end{enumerate}
\section{Fast Information-theoretic Bayesian Optimisation}

Information-theoretic techniques aim to reduce the uncertainty about the unknown global minimiser $\mathbf{x}_*$ by selecting a query point that leads to the largest reduction in entropy of the distribution $p(\mathbf{x}_* \vert D_n)$  \citep{hennig2012entropy}.  The acquisition function for such techniques has the form \citep{{hennig2012entropy},{hernandez2014predictive}}:
\begin{align} \label{ESacfunc}
	\alpha_{\scaleto{\text{ES}\mathstrut}{5pt}} (\mathbf{x} \vert D_n ) &= H[p(\mathbf{x}_{*} \vert D_n)] \nonumber
	\\&- \ \mathbb{E} _{p(y \vert D_n,\mathbf{x})}\Big[H \big[p \big(\mathbf{x}_{*} \vert D_n \cup {(\mathbf{x},y)} \big) \big] \Big].
\end{align}
\\	
PES makes use of the symmetry of mutual information and arrives at the following equivalent acquisition function:		\begin{align} \label{PESacfunc}
	\alpha_{\scaleto{\text{PES}\mathstrut}{5pt}} (\mathbf{x} \vert D_n ) &= H[p(y \vert D_n,\mathbf{x})]  \nonumber
	\\ & - \mathbb{E} _{p(\mathbf{x}_{*} \vert D_n)}\Big[H \big[p(y \vert D_n,\mathbf{x},\mathbf{x}_{*}) \big] \Big],
\end{align}
	
where $p(y \vert D_n, \mathbf{x}, \mathbf{x}_{*})$ is the predictive posterior distribution for $y$ conditioned on the observed data $D_n$, the test location $\mathbf{x}$ and the global minimiser $ \mathbf{x}_{*}$ of the objective function.      
\\\\ 
FITBO harnesses the same information-theoretic thinking but measures the entropy about the latent global minimum $f_{*}=f(\mathbf{x}_{*})$ instead of that of the global minimiser $\mathbf{x}_{*}$. Thus, the acquisition function of FITBO method is the mutual information between the function minimum $f_{*}$ and the next query point \citep{wang2017max}. In other words, FITBO aims to select the next query point which minimises the entropy of the global minimum: 
	\begin{align} \label{FITBOacfunc}
	\alpha_{\scaleto{\text{FITBO}\mathstrut}{5pt}}(\mathbf{x} \vert D_n ) &= H[p(y \vert D_n,\mathbf{x})]  \nonumber
	\\&- \mathbb{E} _{p(f_{*} \vert D_n)}\Big[H \big[p(y \vert D_n,\mathbf{x},f_{*})\big]\Big]. 
	\end{align}        
\\	
This idea of changing entropy computation from the input space to the output space is also shared by \citet{hoffman:2015} and \citet{wang2017max}. Hence, the acquisition function of the FITBO method is very similar to those of OPES \citep{hoffman:2015}  and MES \citep{wang2017max}. 
\\\\
However, our novel contribution is to express the unknown objective function in a parabolic form $f(\mathbf{x}) =\eta+\sfrac{1}{2}g(\mathbf{x})^2$, thus representing the global minimum $ f_{*} $ by a hyperparameter  $ \eta $ and circumventing the laborious process of sampling the global minimum. FITBO acquisition function can then be reformulated as: 
   	\begin{align} \label{FITBOacfunc2}
	\alpha_{\scaleto{\text{FITBO}\mathstrut}{5pt}}(\mathbf{x} \vert D_n ) &= H[p(y \vert D_n,\mathbf{x})]  \nonumber
	\\ &- \mathbb{E} _{p(\eta \vert D_n)}\Big[H \big[p(y \vert D_n,\mathbf{x},\eta) \big] \Big] \nonumber
	\\&= H \Big[ \int p(y \vert D_n,\mathbf{x},\eta) p(\eta \vert D_n) \mathrm{d}\eta \Big] \nonumber
	\\&- \int p(\eta \vert D_n) H \big[p(y \vert D_n,\mathbf{x},\eta) \big] \mathrm{d}\eta. 
	\end{align}  
\\
The intractable integral terms can be approximated by drawing $M$ samples of $\eta$ from the posterior distribution $p(\eta \vert D_n)$ and using a Monte Carlo method \citep{hernandez2014predictive}. The predictive posterior distribution $p(y \vert D_n,\mathbf{x}, \eta)$ can be turned into a neat Gaussian form by applying a local linearisation technique on our parabolic transformation as described in Section \ref{prarabolicandlinearisation}. Thus, the first term in the above FITBO acquisition function is an entropy of a Gaussian mixture, which is intractable and demands approximation as described in Section \ref{approxientropyofGMM}. The second term is the expected entropy of a one-dimensional Gaussian distribution and can be computed analytically because the entropy of a Gaussian has the closed form: $H[p(y \vert D_n,\mathbf{x},\eta)] = 0.5 \log \big[2\pi e \big(v_f(\mathbf{x} \vert D_n,\eta )+\sigma_n^2 \big) \big] $ where the variance $v_f(\mathbf{x} \vert D_n,\eta )=K_f(\mathbf{x},\mathbf{x}')$ and $\sigma_n^2$ is the variance of observation noise.

\subsection{Parabolic Transformation and Predictive Posterior Distribution} \label{prarabolicandlinearisation}

\citet{gunter2014sampling} use a square-root transformation on the integrand in their warped sequential active Bayesian integration method to ensure non-negativity. Inspired by this work, we creatively express any unknown objective function $f(\mathbf{x}) $ in the parabolic form:
	\begin{equation} 
     f(\mathbf{x}) =\eta+\sfrac{1}{2}g(\mathbf{x})^2 ,
	\end{equation} 
where $\eta$ is the global minimum of the objective function. Given the noise-free observation data $D_f=\{(\mathbf{x_{i}},f_{i}) \vert i=1, \dots n\}=\{ \mathbf{X}_n,\mathbf{f}_n\}$, the observation data on $g$ is $D_g=\{(\mathbf{x_{i}},g_{i}) \vert i=1, \dots n\}=\{ \mathbf{X}_n,\mathbf{g}_n\}$ where $g_{i}=\sqrt{ 2(f_i-\eta)}$ . 
\\\\
We impose a zero-mean Gaussian process prior on $g(\mathbf{x})$, $ g \sim \mathcal{GP} \big(0, k(\mathbf{x},\mathbf{x}') \big)$, so that the posterior distribution for $ g $ conditioned on the observation data $D_g$ and the test point $\mathbf{x} $ also follows a Gaussian process:
\begin{equation}
p(g \vert D_g,\mathbf{x} ,\eta) = \mathcal{GP} \big(g; m_{g} (\cdot), K_{g} (\cdot,\cdot) \big) 
\end{equation}
where \[ m_g (\mathbf{x}) = K(\mathbf{x}, \mathbf{X}_n)  K(\mathbf{X}_n,\mathbf{X}_n)^{-1} \mathbf{g}_n, \] 
\[K_g (\mathbf{x},\mathbf{x}')=K(\mathbf{x},\mathbf{x}') - K( \mathbf{x}, \mathbf{X}_n) K(\mathbf{X}_n,\mathbf{X}_n)^{-1} K(\mathbf{X}_n,\mathbf{x}').\]
\\\noindent
The parabolic transformation causes the distribution for any $f $ to become a non-central $\chi^2$ process, making the analysis intractable. In order to tackle this problem and obtain a posterior distribution $ p(f \vert D_f, \mathbf{x},\eta)$ that is also Gaussian, we employ a linearisation technique \citep{gunter2014sampling}.
\\\\ \noindent
We perform a local linearisation of the parabolic transformation $ h(g) =\eta+\sfrac{1}{2}g^2$ around $g_0$ and obtain $f \approx h(g_0)+h'(g_0)(g - g_0)$ where the gradient $h'(g)=g$. By setting $ g_0 $ to the mode of the posterior distribution  $p(g \vert D_g,\mathbf{x}, \eta )$ (i.e. $ g_0 = m_g $), we obtain an expression for $f$ which is linear in $g$:
      \begin{align} 
	f(\mathbf{x}) &\approx [\eta+\sfrac{1}{2}m_g(\mathbf{x})^2 ] + m_g(\mathbf{x}) [ g(\mathbf{x}) - m_g(\mathbf{x}) ] \nonumber
	\\& = \eta - \sfrac{1}{2}m_g(\mathbf{x})^2 + m_g(\mathbf{x}) g(\mathbf{x}).
        \end{align}
\noindent
Since the affine transformation of a Gaussian process remains Gaussian, the predictive posterior distribution for $f$ now has a closed form:
	\begin{equation} \label{posteriorf}
	p(f \vert D_f,\mathbf{x},\eta) = \mathcal{GP} \big(f; m_{f} (\cdot), K_{f} (\cdot,\cdot) \big)   
	\end{equation}   
where
	 \[ m_f (\mathbf{x}) = \eta + \sfrac{1}{2}m_g(\mathbf{x})^2 \] 
	\[ K_f (\mathbf{x},\mathbf{x}') = m_g(\mathbf{x})K_g (\mathbf{x},\mathbf{x}') m_g(\mathbf{x}'). \]
\\
However, in real world situations, we do not have access to the true function values but only noisy observations of the function, $ y(\mathbf{x}) = f(\mathbf{x}) +  \epsilon $, where $\epsilon$ is assumed to be an independently and identically distributed Gaussian noise with variance $\sigma_n^2$ \citep{rasmussen2006gaussian}. Given noisy observation data $D_n=\{(\mathbf{x_{i}},y_{i}) \vert i=1, \dots n\}=\{ \mathbf{X}_n,\mathbf{y}_n\}$, the predictive posterior distribution (\ref{posteriorf}) becomes:
 	\begin{equation} \label{posteriory}
	p(y \vert D_n,\mathbf{x},\eta )=\mathcal{GP} \big(y; m_{f} (\cdot), K_{f} (\cdot,\cdot)+\sigma_n^2 \delta(\cdot, \cdot) \big). 
 	\end{equation}

\subsection{Hyperparameter Treatment}
Hyperparameters are the free parameters, such as output scale and characteristic length scales in the kernel function for the Gaussian processes as well as noise variance. We use $\boldsymbol{\theta}$ to represent a vector of hyperparameters that includes all the kernel parameters and the noise variance. Recall that we introduce a new hyperparameter $\eta$ in our model to represent the global minimum. 
To ensure that $\eta$ is not greater than the minimum observation $y_{min}$, we assume that $ \log (y_{min}- \eta) $ follows a broad normal distribution. Thus the prior for $\eta$ has the form: 
\begin{equation}
	p(\eta) = \frac{1} { (y_{min}- \eta)} \mathcal{N} \big( \log (y_{min}- \eta); \mu , \sigma^2 \big).	
\end{equation}

The most popular approach to hyperparameter treatment is to learn hyperparameter values via maximum likelihood estimation (MLE) or maximum a posterior estimation (MAP). However, both MLE and MAP are not desirable as they give point estimates and ignore our uncertainty about the hyperparameters \citep{hernandez2014predictive}. In a fully Bayesian treatment of the hyperparameters, we should consider all possible hyperparameter values. This can be done by marginalising the terms in the acquisition function with respect to the posterior  $p(\boldsymbol{\psi} \vert D_n )$ where $ \boldsymbol{\psi} = \{\boldsymbol{\theta}, \eta \}$:
	\begin{align*} \label{FITBOacfunc3}
	\alpha_{\scaleto{\text{FITBO}\mathstrut}{5pt}}(\mathbf{x} \vert D_n ) & = H \Big[ \int p(y \vert D_n,\mathbf{x},\boldsymbol{\psi}) p(\boldsymbol{\psi} \vert D_n) \mathrm{d}\boldsymbol{\psi} \Big] \nonumber
	\\&- \int p(\boldsymbol{\psi} \vert D_n) H\big[p(y \vert D_n,\mathbf{x},\boldsymbol{\psi}) \big] \mathrm{d}\boldsymbol{\psi}. 
	\end{align*}  

Since complete marginalisation over hyperparameters is analytically intractable, the integral can be approximated using the Monte Carlo method \citep{{hoffman:2015},{snoek2012practical}}, leading to the final expression:
	  \begin{align}
&\alpha_{\scaleto{\text{FITBO}\mathstrut}{5pt}}(\mathbf{x} \vert D_n) \nonumber
	\\&= H \Big[  \frac{1}{M} \sum_j^{M} p(y \vert D_n,\mathbf{x},\boldsymbol{\theta}^{(j)}, \eta^{(j)}) \Big]  \nonumber
	\\& \quad-  \frac{1}{2M} \sum_j^{M} \log \big[ 2\pi e \big(v_f(\mathbf{x} \vert D, \boldsymbol{\theta}^{(j)}, \eta^{(j)})+\sigma_n^2 \big) \big]. 
   	  \end{align}

\subsection{Approximation for the Gaussian Mixture Entropy}
\label{approxientropyofGMM}
The entropy of a Gaussian mixture is intractable and can be estimated via a number of methods: the Taylor expansion proposed in \citep{huber2008entropy}, numerical integration and Monte Carlo integration. Of these three, our experimentation revealed that numerical integration (in particular, an adaptive Simpson's method) was clearly the most performant for our application (see the supplementary material). Note that our Gaussian mixture is univariate.
\\\\
A faster alternative is to approximate the first entropy term by matching the first two moments of a Gaussian mixture. The mean and variance of a univariate Gaussian mixture model $p (z) = \sum_j^{M}  \frac{1}{M}  \mathcal{N}(z \vert m_j, K_j)$ have the analytical form:
	\begin{equation}
	\mathbb{E} [z] = \sum_j^{M}  \frac{1}{M}  m_j 
	\end{equation}
	\begin{equation}
	\text{Var}(z)  =  \sum_j^{M}  \frac{1}{M} ( K_j + m_j^2) - \mathbb{E} [z] ^2.
	\end{equation}
By fitting a Gaussian to the Gaussian mixture, we can obtain a closed-form upper bound for the first entropy term: $ H[p(z)] \approx 0.5 \log \big[2\pi e \big( \text{Var}(z)+\sigma_n^2 \big) \big] $, thus further enhancing the computational speed of FITBO approach. However, the moment-matching approach results in a looser approximation than numerical integration (shown in the supplementary material) and we will compare both approaches in our experiments in Section \ref{sec:Experiments}. 

\subsection{The Algorithm}

\begin{figure} [h!btp]  
        	\begin{subfigure}{1.0\linewidth}
        		  \centering
        		  \includegraphics[trim=0.5cm 0.2cm 0.5cm  0.2cm, clip, width=1.0\linewidth]{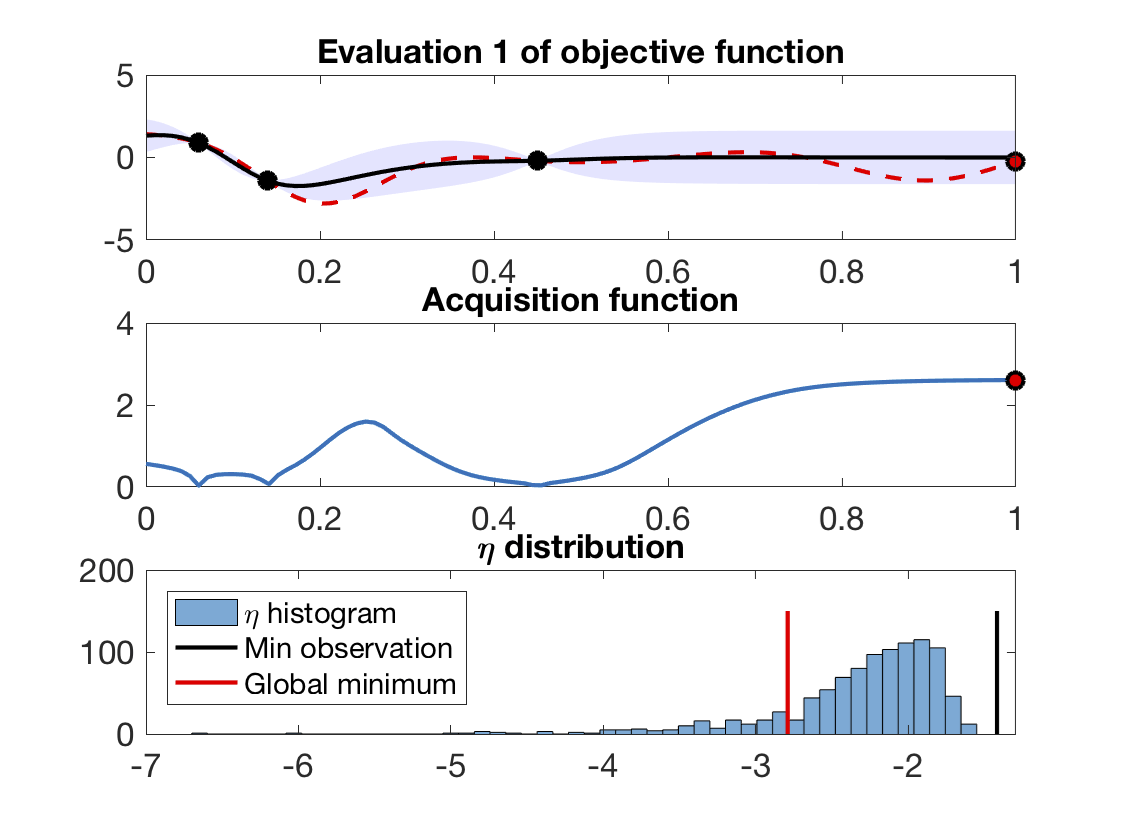}
        	\end{subfigure}
        	\begin{subfigure}{1.0\linewidth}
        		  \centering
        		  \includegraphics[trim=0.5cm 0.2cm 0.5cm  0.2cm, clip, width=1.0\linewidth]{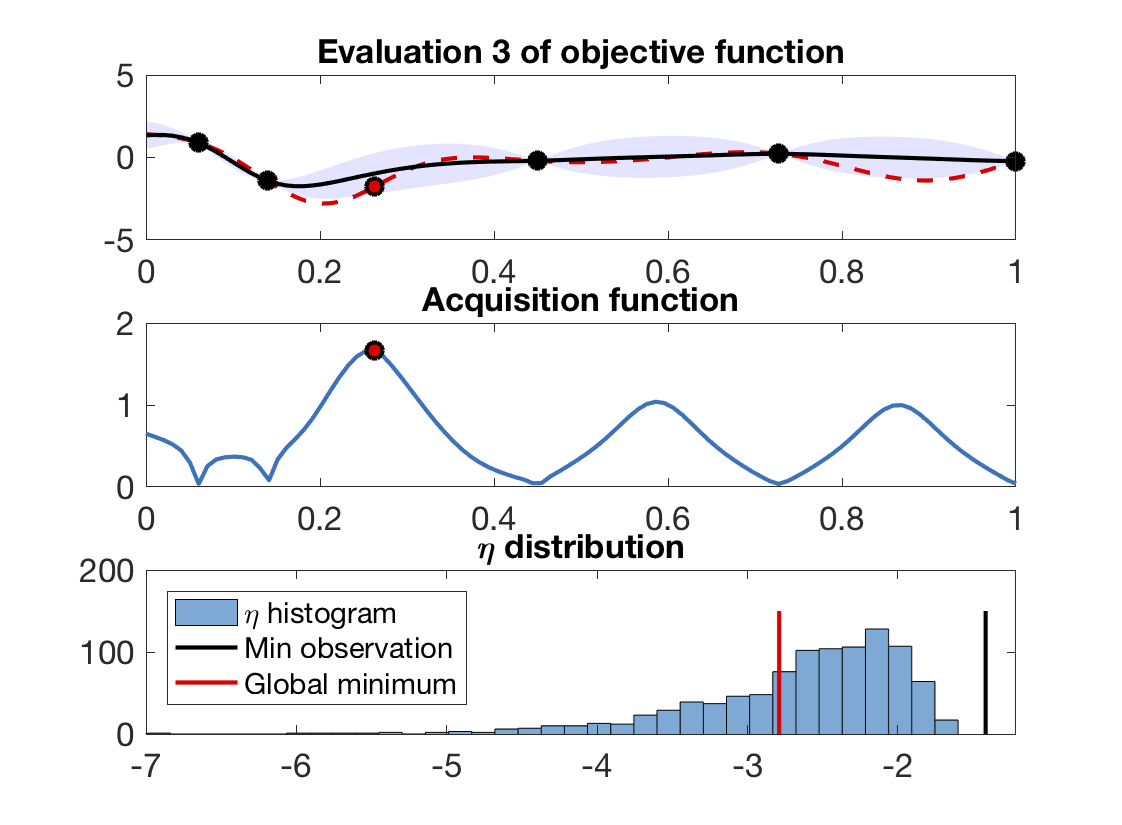}
        	\end{subfigure}
        	\begin{subfigure}{1.0\linewidth}
        		 \centering
        		 \includegraphics[trim=0.5cm 0.2cm 0.5cm  0.2cm, clip, width=1.0\linewidth]{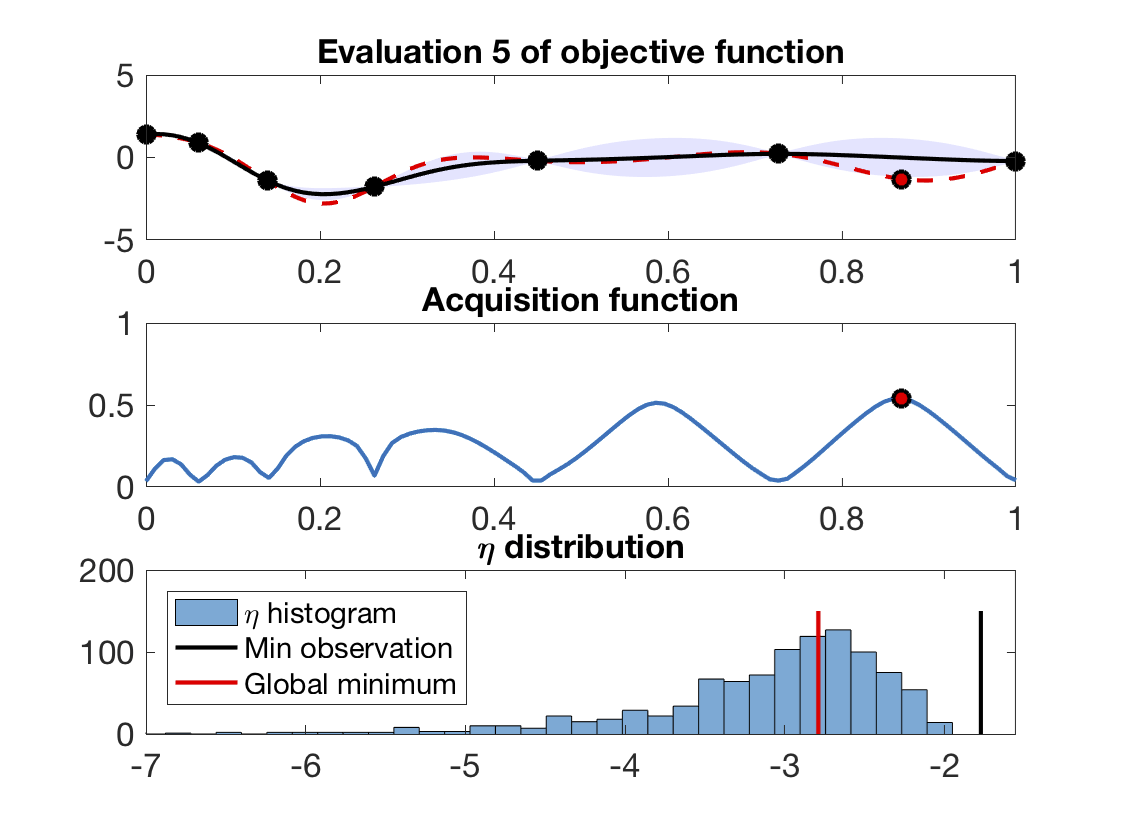}
        	\end{subfigure}	
   	
        \caption{ Bayesian optimisation for a 1D objective function using FITBO method at the 1st, 3rd, 5th evaluations. In each subfigure, the top plot shows the objective function (red dotted line), the posterior mean (black solid line) and the 95$\%$ confidence interval (blue shaded area) estimated by the Gaussian process model as well as the observation points (black dot) and the next query point (red dot). The middle plot shows the acquisition function. The bottom plot is the histogram of $\eta$ samples as well as its relation to the minimum observation (black vertical line) and the true global minimum (red vertical line).}
	\label{1DFITBO}
\end{figure}

The procedures of computing the acquisition function of FITBO are summarised by Algorithm \ref{alg:FITBO}. Figure \ref{1DFITBO} illustrates the sampling behaviour of FITBO method for a simple 1D Bayesian optimisation problem. The optimisation process is started with 3 initial observation data. As more samples are taken, the mean of the posterior distribution for the objective function gradually resembles the objective function and the distribution of $\eta$ converges to the global minimum.

\begin{algorithm} [h]
\caption{FITBO acquisition function}\label{alg:FITBO}
\begin{algorithmic}[1]
       \STATE {\bfseries Input:} a test input $\mathbf{x}$; noisy observation data 
       \\ $D_n=\{(\mathbf{x}_{i},y_{i}) \vert i=1, \dots, n\}$

	\STATE Sample hyperparameters and $\eta$ from $p(\boldsymbol{\psi} \vert D_n)$: 
	\\ $\boldsymbol{\Psi}= \{\boldsymbol{\theta}^{(j)},\eta^{(j)} \vert j=1, \dots, M\}$ 
	\FOR {j = 1, $\dots, M$}
	\STATE Use $f(\mathbf{x}) =\eta+\sfrac{1}{2}g(\mathbf{x})^2 $ to approximate $p(f \vert D_n,\mathbf{x},\boldsymbol{\theta}^{(j)}, \eta^{(j)})= \mathcal{GP} \big(m_f(\cdot),K_f(\cdot,\cdot)\big ) $
	
%
%
	
	\STATE Compute $p(y \vert D_n,\mathbf{x},\boldsymbol{\theta}^{(j)}, \eta^{(j)}) $
	
	\STATE Compute $ H[p(y \vert D_n,\mathbf{x},\boldsymbol{\theta}^{(j)}, \eta^{(j)})]  $
	
	\ENDFOR	
			
	\STATE Estimate the entropy of the Gaussian mixture : 
	
	$ \text{E}_1(\mathbf{x} \vert D_n) = H \Big[  \frac{1}{M} \sum_j^{M} p(y \vert D_n,\mathbf{x},\boldsymbol{\theta}^{(j)}, \eta^{(j)}) \Big] $
	
	\STATE Compute the entropy expectation: 
	\\ $ \text{E}_2(\mathbf{x} \vert D_n)= \frac{1}{M} \sum_j^{M} H[p(y \vert D_n,\mathbf{x},\boldsymbol{\theta}^{(j)}, \eta^{(j)})]
	= \frac{1}{2M} \sum_j^{M} \log \big[ 2\pi e \big(v_f(\mathbf{x} \vert D_n, \boldsymbol{\theta}^{(j)}, \eta^{(j)})+\sigma_n^2\big) \big] $ 	
	
	\STATE \textbf{return}  $\alpha_n(\mathbf{x} \vert D_n)=\text{E}_1(\mathbf{x} \vert D_n) - \text{E}_2(\mathbf{x} \vert D_n)$
	
\end{algorithmic}
\end{algorithm}

\section{Experiments}
\label{sec:Experiments}

We conduct a series of experiments to test the empirical performance of FITBO and compare it with other popular acquisition functions. In this section, FITBO denotes the version using numerical integration to estimate the entropy of the Gaussian mixture while FITBO-MM denotes the version using moment matching. In all experiments, we adopt a zero-mean Gaussian process prior with the squared exponential kernel function and use the elliptical slice sampler \citep{murray2010elliptical} for sampling hyperparameters $\boldsymbol{\theta}$ and $\eta$. For the implementation of EI, PI, GP-UCB, MES and PES, we use the open source Matlab code by \citet{wang2017max} and \citet{hernandez2014predictive}. Our Matlab code for FITBO will be available at \url{https://github.com/rubinxin/FITBO}. We use the type of MES method that samples the global minimum $f(\mathbf{x}_{*})$ from an approximated posterior function $\tilde{f}(\mathbf{x}) = \boldsymbol{\phi} (\mathbf{x})^ {\mathrm{T}} \tilde{\mathbf{a}} $ where $\boldsymbol{\phi} (\mathbf{x})$ is an $m$-dimensional feature vector and $\tilde{\mathbf{a}}$ is a Gaussian weight vector \citep{wang2017max}. This is also the minimiser sampling strategy adopted by PES \citep{hernandez2014predictive}. The computational complexity of sampling $\tilde{\mathbf{a}}$ from its posterior distribution $p(\tilde{\mathbf{a}} \vert D_n)$ is $\mathcal{O}(n^2m)$ when $n<m$ \citep{hernandez2014predictive}. Minimising $\tilde{f}(\mathbf{x})$ to within $\zeta $ accuracy using any grid search or branch and bound optimiser requires $ \mathcal{O} (\zeta^{-d})$ calls to $\tilde{f}(\mathbf{x})$ for $d$-dimensional input data \citep{kandasamy2015high}. For both PES and MES, we apply their fastest versions which draw only 1 minimum or minimiser sample to estimate the acquisition function. 

\subsection{Runtime Tests}
\label{sec:runtimetests}

The first set of experiments measure and compare the runtime of evaluating the acquisition functions $\alpha_n(\mathbf{x} \vert D_n)$ for methods including GP-UCB, PI, EI, PES, MES,  FITBO and FITBO-MM. All the timing tests were performed exclusively on a 2.3 GHz Intel Core i5. The runtime measured excludes the time taken for sampling hyperparameters as well as optimising the acquisition functions. The methodology of the tests can be summarised as follows:
\begin{enumerate}
	\item Generate 10 initial observation data from a $d$-dimensional test function and sample a set of $M$ hyperparameters $\boldsymbol{\Psi}= \{\boldsymbol{\theta}^{(j)},\eta^{(j)} \vert j=1, \dots, M\}$ from the log posterior distribution $\log \tilde p(\boldsymbol{\psi} \vert D_n) $ using the elliptical slice sampler.
	\item Use this set of hyperparameters to evaluate all acquisition functions at 100 test points.
	\item Repeat the procedures 1 and 2 for 100 different initialisations and compute the mean and standard deviation of the runtime taken for evaluating various acquisition functions. 
\end{enumerate}

We did not include the time for sampling $\eta$ alone into the runtime of evaluating FITBO and FITBO-MM because $\eta$ is sampled jointly with other hyperparameters and does not add to the overall sampling burden significantly. In fact, we have tested that sampling $\eta$ by the elliptical slice sampler adds 0.09 seconds on average when drawing 2\,000 samples and 0.93 seconds when drawing 20\,000 samples. Note further that we will limit all methods to a fixed number of hyperparameter samples in both runtime tests and performance experiments: this will impart a slight performance penalty to our method, which must sample from a hyperparameter space of one additional dimension. 
\\
\begin{figure} [t] 
        		  \centering
        		  \includegraphics[trim=0.5cm 0.2cm 0.7cm  0.2cm, clip, width=1.0\linewidth]{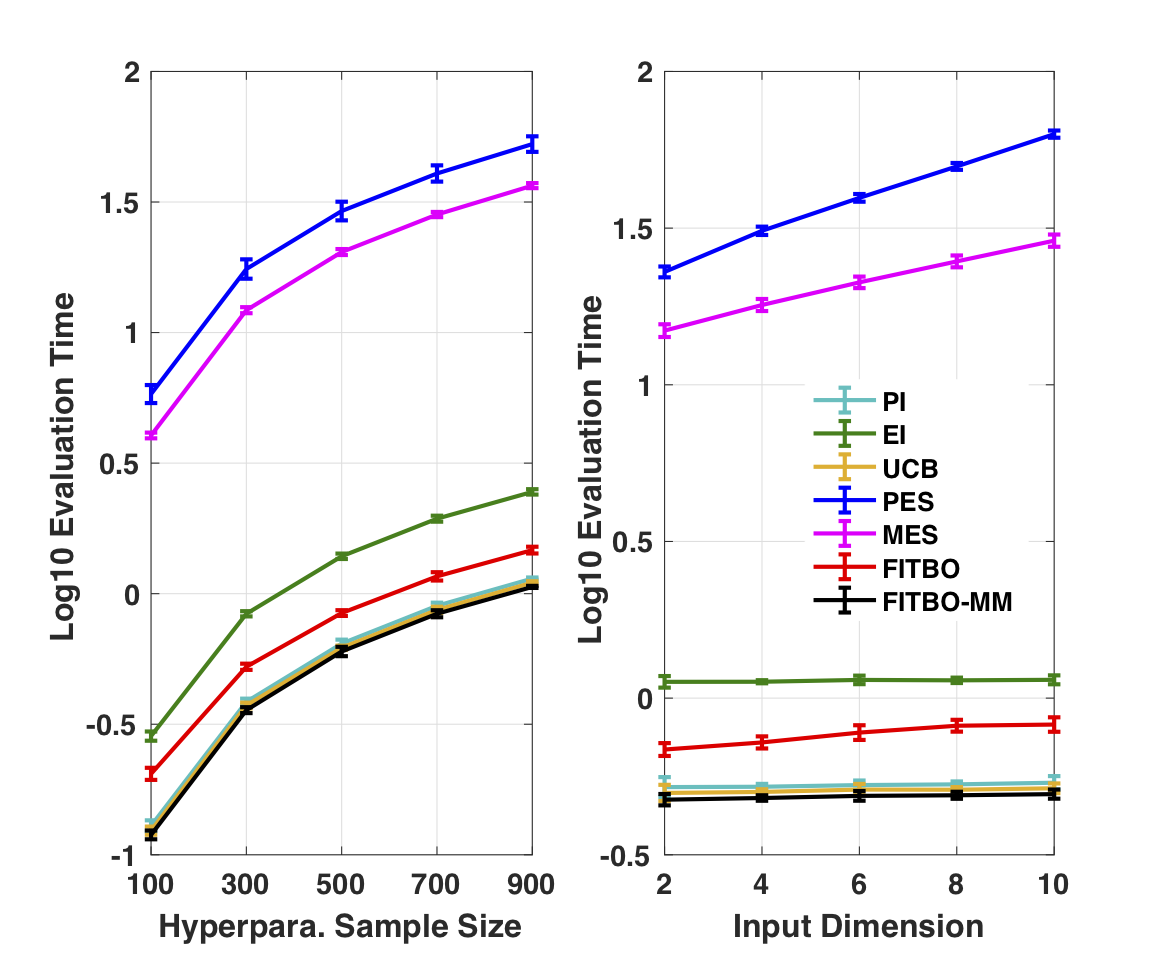}
		   \caption{ The runtime of evaluating 7 different acquisition functions (PI, EI, GP-UCB, PES, MES, FITBO and FITBO-MM) at 100 test inputs. The left plot shows the runtime of evaluating the acquisition functions using $M$ hyperparameter samples for 2D input data and $M$ tested are 100, 300, 500, 700, 900. The right plot shows the runtime of evaluating the acquisition functions using 400 hyperparameter samples for input data of dimension $d$ where $d$ are 2, 4, 6, 8, 10. The y-axes are the evaluation runtime expressed in the logarithm to the base 10.}
		   \label{runtimeacrossdimensions}
\end{figure}

\begin{table} [h]
\caption{Runtime of evaluating PI, GP-UCB and FITBO-MM at 100 2D inputs using $M$ hyperparameter samples (Unit: Second).} \label{diffsamplesize}
\begin{center} 
\begin{tabular} {cccc} \toprule \addlinespace
{\bf{M}} &{\bf PI} &{\bf GP-UCB} &{\bf FITBO-MM} \\ \midrule \addlinespace
100           &  0.1293          &  0.1238 		&  \bf{0.1193}  \\
           & ($\pm$ 0.006)	& ($\pm$ 0.005) 	&($\pm$ 0.005)  \\\addlinespace
300            & 0.3856     	&  0.3731  		&   \bf{0.3582} \\
           & ($\pm$ 0.011)	& ($\pm$ 0.010) 	&($\pm$ 0.009)  \\\addlinespace
500             & 0.6442  	 &  0.6205  	&   \bf{0.6011} \\
           & ($\pm$ 0.025)	 & ($\pm$ 0.012) 	&($\pm$ 0.027)  \\\addlinespace
700           & 0.8990             &  0.8670  		&   \bf{0.8382} \\
           & ($\pm$ 0.026)	  & ($\pm$ 0.026) 	&($\pm$ 0.028)  \\\addlinespace
900            & 1.1426   	  &  1.1025 	&   \bf{1.0618} \\
           & ($\pm$ 0.011)	& ($\pm$ 0.014) 	&($\pm$ 0.010)  \\\addlinespace
           \bottomrule
\end{tabular}
\end{center}
\end{table}

\begin{table}[h]
\caption{Runtime of evaluating PI, GP-UCB and FITBO-MM for 100 test inputs of dimension $d$ with M = 400 (Unit: Second).} \label{dimension}
\begin{center}
\begin{tabular} {cccc} \toprule \addlinespace
\bf {d}  &{\bf PI} 		&{\bf GP-UCB} 		&{\bf FITBO-MM} \\ \midrule \addlinespace
2       &0.5217                &  0.4991 			&   \bf{0.4745} \\ 
           & ($\pm$ 0.047)	& ($\pm$ 0.034) 	&($\pm$ 0.021)  \\\addlinespace
4      &0.5215  	    		&  0.5020 			&   \bf{0.4800}\\
           & ($\pm$ 0.011)	& ($\pm$ 0.010) 	&($\pm$ 0.010)  \\\addlinespace
6       &0.5281 	    		& 0.5112 			&    \bf{0.4879}\\
           & ($\pm$ 0.019)	& ($\pm$ 0.023) 	&($\pm$ 0.019)  \\\addlinespace
8       &0.5307  	    	& 0.5102 			&    \bf{0.4899}\\
           & ($\pm$0.011)	& ($\pm$ 0.010) 	&($\pm$ 0.013)  \\\addlinespace
10     &0.5378 	    		& 0.5159 			&    \bf{0.4942}\\
           & ($\pm$ 0.029)	& ($\pm$ 0.019) 	&($\pm$ 0.017)  \\\addlinespace
           
           \bottomrule
\end{tabular}
\end{center}
\end{table}

The above tests are repeated for different hyperparameter sample sizes $M=100, 300, 500, 700, 900$ and input data of different dimensions $d=2, 4, 6, 8, 10$. The results are presented graphically in Figure \ref{runtimeacrossdimensions} with the evaluation runtime being expressed in the logarithm to the base 10 and the exact numerical results for methods that are very close in runtime are presented in Tables \ref{diffsamplesize} and \ref{dimension}.  
\\\\
Figure \ref{runtimeacrossdimensions} shows that FITBO is significantly faster to evaluate than PES and MES for various hyperparameter sample sizes used and for problems of different input dimensions. Moreover, FITBO even gains a clear speed advantage over EI. The moment matching technique manages to further enhance the speed of FITBO, making FITBO-MM comparable with, if not slightly faster than, simple algorithms like PI and GP-UCB. In addition, we notice that the runtime of evaluating FITBO-MM, EI, PI and GP-UCB tend to remain constant regardless of the input dimensions while the runtime for PES and MES tends to increase with input dimensions at a rate of $10^{d}$. Thus, our approach is more efficient and applicable in dealing with high-dimensional problems.

\subsection{Tests with Benchmark Functions}

We perform optimisation tasks on three challenging benchmark functions: Branin (defined in $[0, 1]^2$), Eggholder (defined in $[0, 1]^2$) and Hartmann (defined in $[0, 1]^6$). In all tests, we set the observation noise to $\sigma_n^2 = 10^{-3}$ and resample all the hyperparameters after each function evaluation. In evaluating the optimisation performance of various Bayesian optimisation methods, we use the two common metrics adopted by \citet {hennig2012entropy}. The first metric is Immediate regret (IR) which is defined as:
	\begin{equation} \label {IR}
		IR=\vert f(\mathbf{x}_{*})- f(\hat{\mathbf{x}}_n)\vert 
	\end{equation}
where $\mathbf{x}_{*}$ is the location of true global minimum and $\hat{\mathbf{x}}_n$ is the best guess recommended by a Bayesian optimiser after $n$ iterations, which corresponds to the minimiser of the posterior mean. The second metric is the Euclidean distance of an optimiser's recommendation $\hat{\mathbf{x}}_n$ from the true global minimiser $\mathbf{x}_{*}$, which is defined as:
	\begin{equation} \label{Euclidean}
		\| L\| _{2} =\| \mathbf{x}_{*} - \hat{\mathbf{x}}_{n} \|. 
	\end{equation} 

We compute the median IR and the median $\| L\| _{2}$ over 40 random initialisations.  At each initialisation, all Bayesian optimisation algorithms start from 3 random observation data for Branin-2D and Eggholder-2D problems and from 9 random observation data for Hartmann-6D problem.

\begin{figure} [h!btp]
	\begin{subfigure}{1.0\linewidth}
	        		  \centering
        		  \includegraphics[trim=0.2cm 0cm 0.3cm  0.1cm, clip, width=1.0\linewidth]{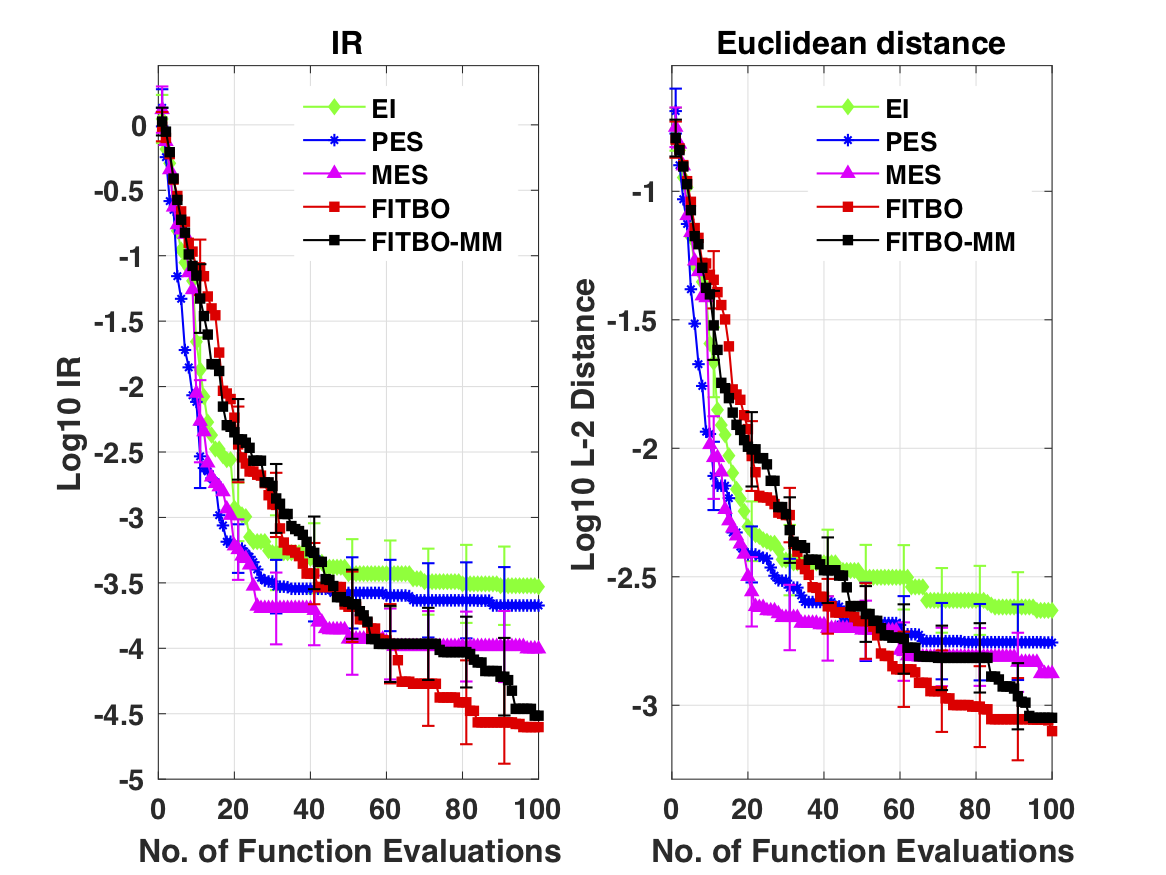}
		   \caption{Branin-2D}
	\end{subfigure}
        \begin{subfigure} {1.0\linewidth}
                		  \centering
                		  \includegraphics[trim=0.2cm 0cm 0.3cm  0.1cm, clip, width=1.0\linewidth]{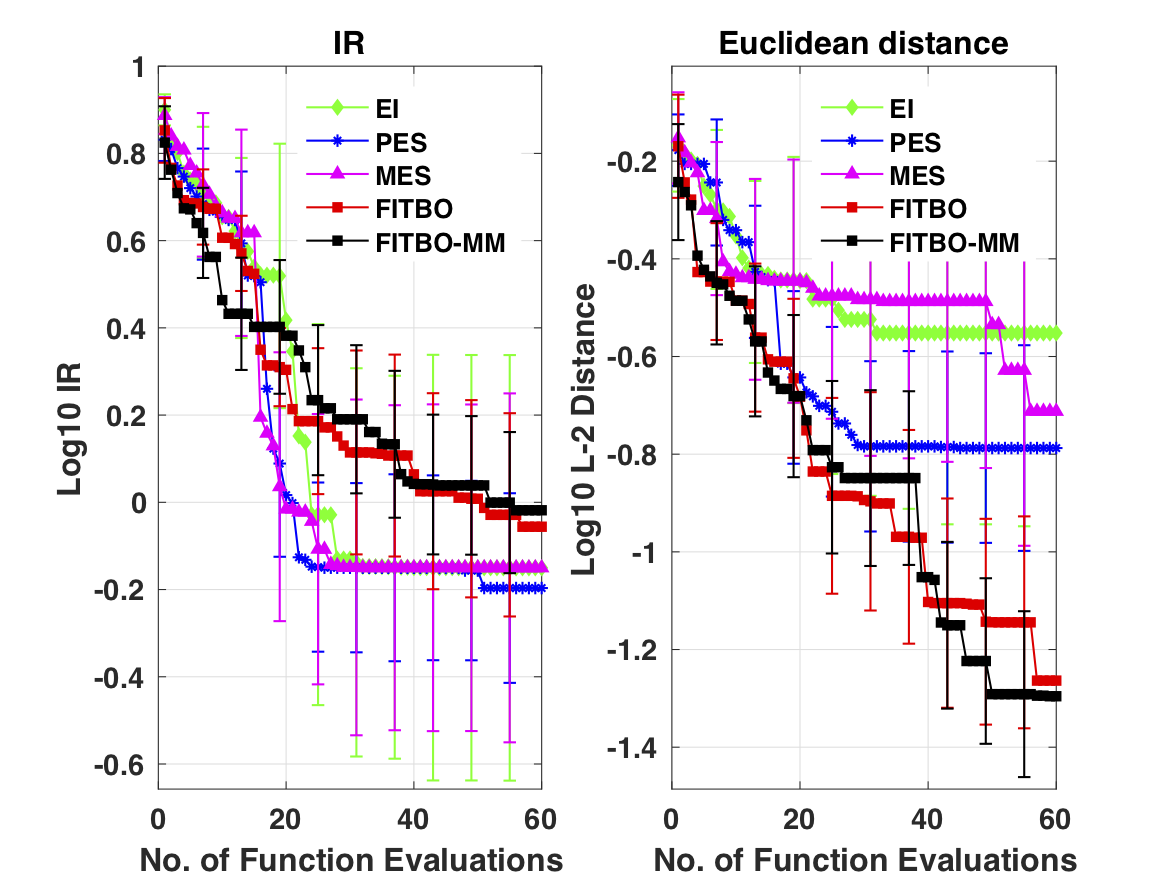}
        		   \caption{Eggholder-2D}
        \end{subfigure}
        \begin{subfigure} {1.0\linewidth}
                		  \centering
                		  \includegraphics[trim=0.2cm 0cm 0.3cm  0.1cm, clip, width=1.0\linewidth]{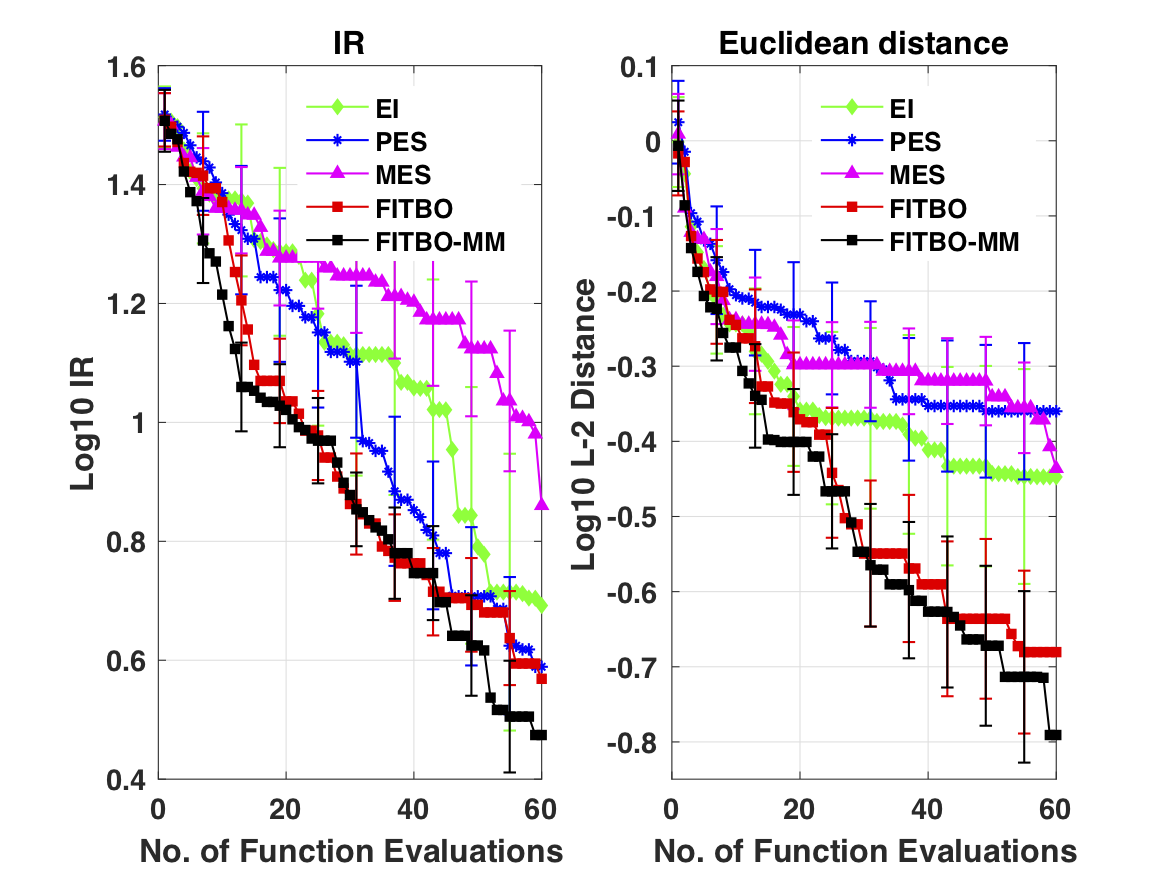}
        		   \caption{Hartmann-6D}
        \end{subfigure}
        \caption{ Optimisation performance of EI, PES, MES, FITBO and FITBO-MM for three benchmark test functions.}
        \label{Benchmarkfunction}
\end{figure}

\begin{figure} [h!btp] 
        		  \centering
        		  \includegraphics[trim=1.1cm 0.8cm 1.1cm  0.1cm, clip, width=0.90\linewidth]{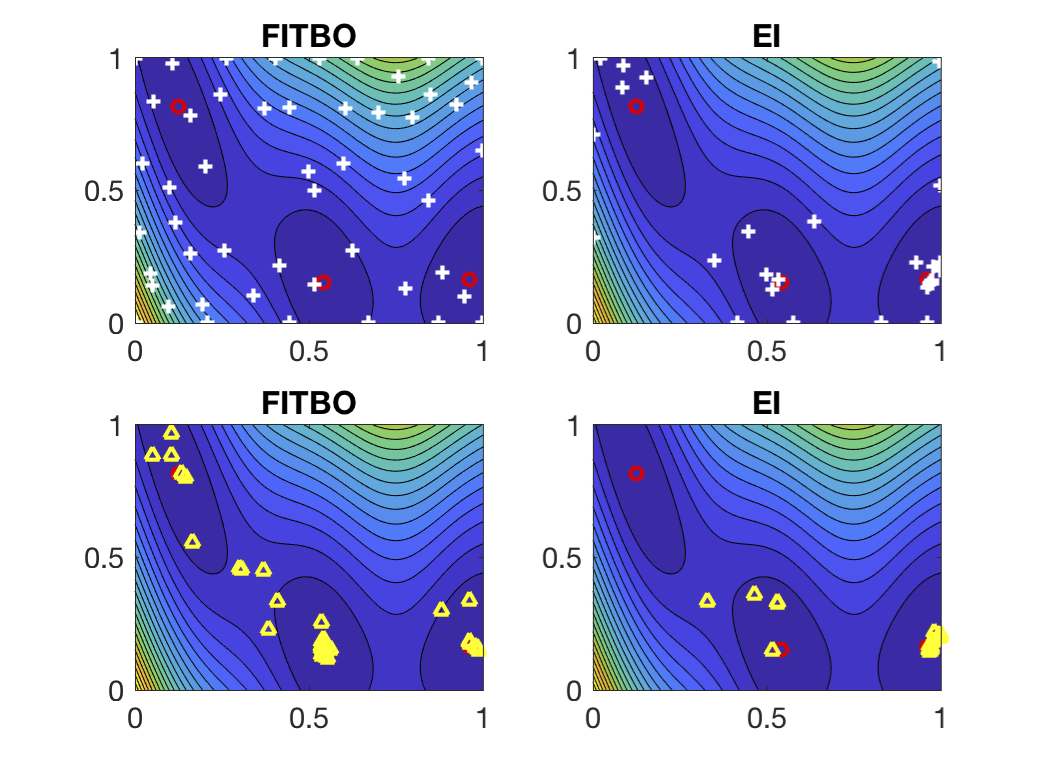}
		   \caption{ Evaluations taken by FITBO and EI in the Branin-2D problem. The white crosses in the top plots indicate the first 50 query points recommended by the two algorithms. The yellow triangles in the bottom plots indicate the guesses of the global minimiser recommended by the algorithms (i.e. $\hat{\mathbf{x}}_n$) after each evaluation. FITBO, which is more explorative in taking evaluations, successfully identifys all three global minimisers (red circle) but EI misses out one of the global minimisers.}	   
		   \label{2Dbraninevaluationpoints}
\end{figure}

\smallbreak
The results are presented in Figure \ref{Benchmarkfunction}. The plots on the left show the median IR achieved by each approach as more evaluation steps are taken. The plots on the right show the median $\| L\| _{2}$ between each optimiser's recommended global minimiser and the true global minimiser. The error bars indicate one standard deviation. 
\\\\
In the case of Branin-2D, FITBO and FITBO-MM lose out to other methods initially but surpass other methods after 50 evaluations. One interesting point we would like to illustrate through the Branin problem is the fundamentally different mechanisms behind information-based approaches like FITBO and improvement-based approaches like EI. As shown in Figure \ref{2Dbraninevaluationpoints}, FITBO is much more explorative compared to EI in taking new evaluations because FITBO selects the query points that maximise the information gain about the minimiser instead of those that lead to an improvement over the best function value observed. FITBO successfully finds all three global minimisers but EI quickly concentrates its searches into regions of low function values, missing out one of the global minimisers.
\\\\
In the case of Eggholder-2D which is more complicated and multimodal, FITBO and FITBO-MM perform not as well as other methods in finding lower function values but outperform all competitors in locating the global minimiser by a large margin. One reason is that the function value near the global minimiser of Eggholder-2D rises sharply. Thus, although FITBO and FITBO-MM are able to better identify the location of the true global minimum, they return higher function values than other methods that are trapped in locations of good local minima. 
\\\\
As for a higher dimensional problem, Hartmann-6D, FITBO and FITBO-MM outperform all other methods in finding both the lower function value and the location of the global minimum. In all three tasks, FITBO-MM, despite using a looser upper bound of the Gaussian mixture entropy, still manages to demonstrate similar, sometimes better, results compared with FITBO. This shows that the performance of our information-theoretic approach is robust to slightly worse approximation of the Gaussian mixture entropy. 

\subsection{Test with Real-world Problems}

Finally, we experiment with a series of real-world optimisation problems.  The first problem (Boston) returns the L2 validation loss of a 1-hidden layer neural network \citep{wang2017max} on the Boston housing dataset \citep{bache2013uci}. The dataset is randomly partitioned into train/validation/test sets and the neural network is trained with Levenberg-Marquardt optimisation. The 2 variables tuned with Bayesian optimisation are the number of neurons and the damping factor $\mu$. 
\\\\
The second problem (MNIST-SVM) outputs the classification error of a support vector machine (SVM) classifier on the validation set of the MNIST dataset \citep{lecun1998gradient}. The SVM classifier adopts a radial basis kernel and the 2 variables to optimise are the kernel scale parameter and the box constraint. 
\\\\
The third problem (Cancer) returns the cross-entropy loss of a 1-hidden layer neural network \citep{wang2017max} on the validation set of the breast cancer dataset \citep{bache2013uci}. This neural network is trained with the scaled conjugate gradient method and we use Bayesian optimisation methods to tune the number of neurons, the damping factor $\mu$, the $\mu-$increase factor and the $\mu-$decrease factor.  

\begin{figure} [t] 
	\begin{subfigure}{0.49\linewidth}
	        		  \centering
        		  \includegraphics[trim=0.5cm 0.1cm 10.75cm  0.1cm, clip, width=1.0\linewidth]{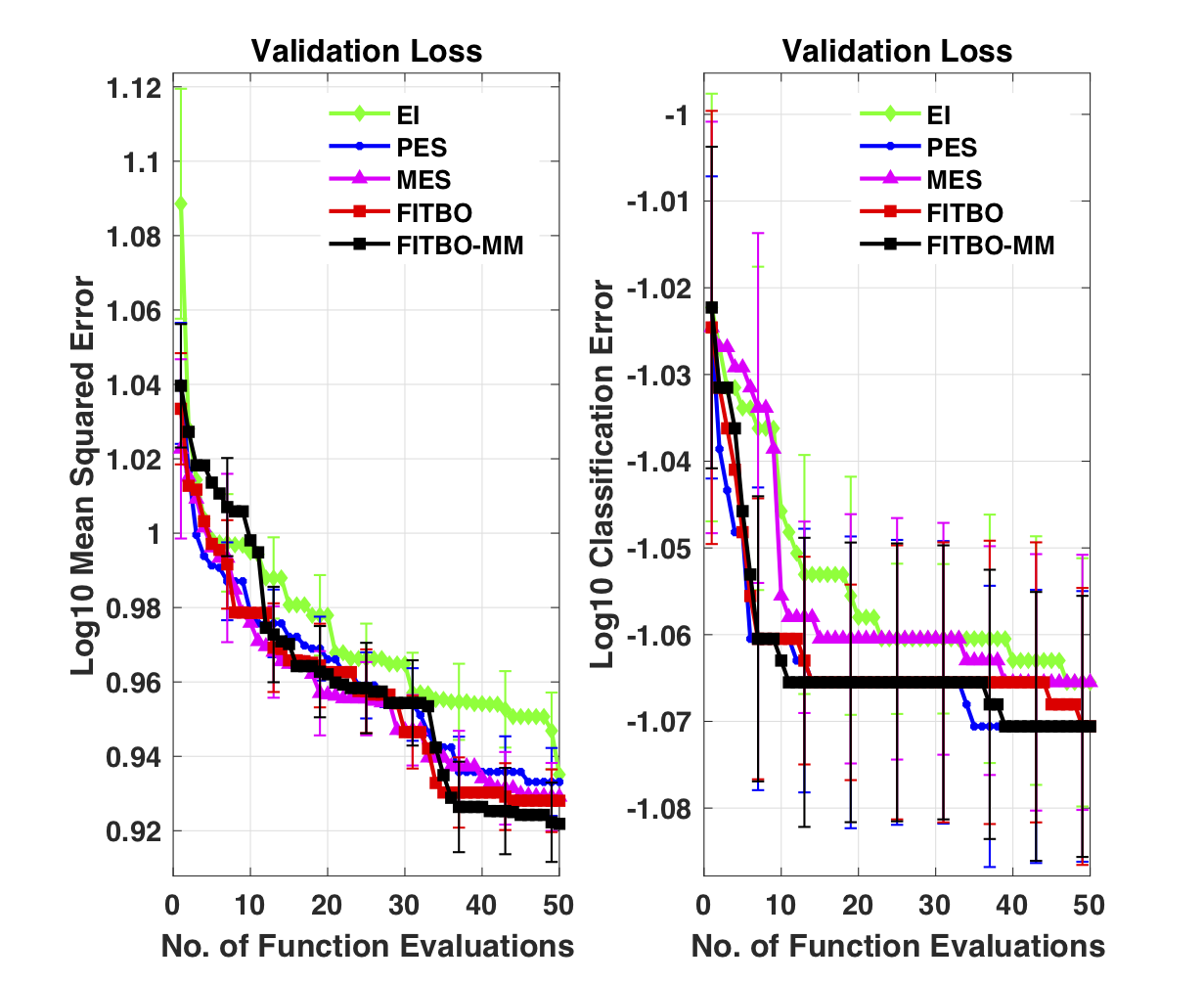}
		   \caption{Boston}
		   \label{Boston}
	\end{subfigure}
        \begin{subfigure} {0.49\linewidth}
                		  \centering
                		  \includegraphics[trim=10.1cm 0.1cm 1.15cm  0.1cm, clip, width=1.0\linewidth]{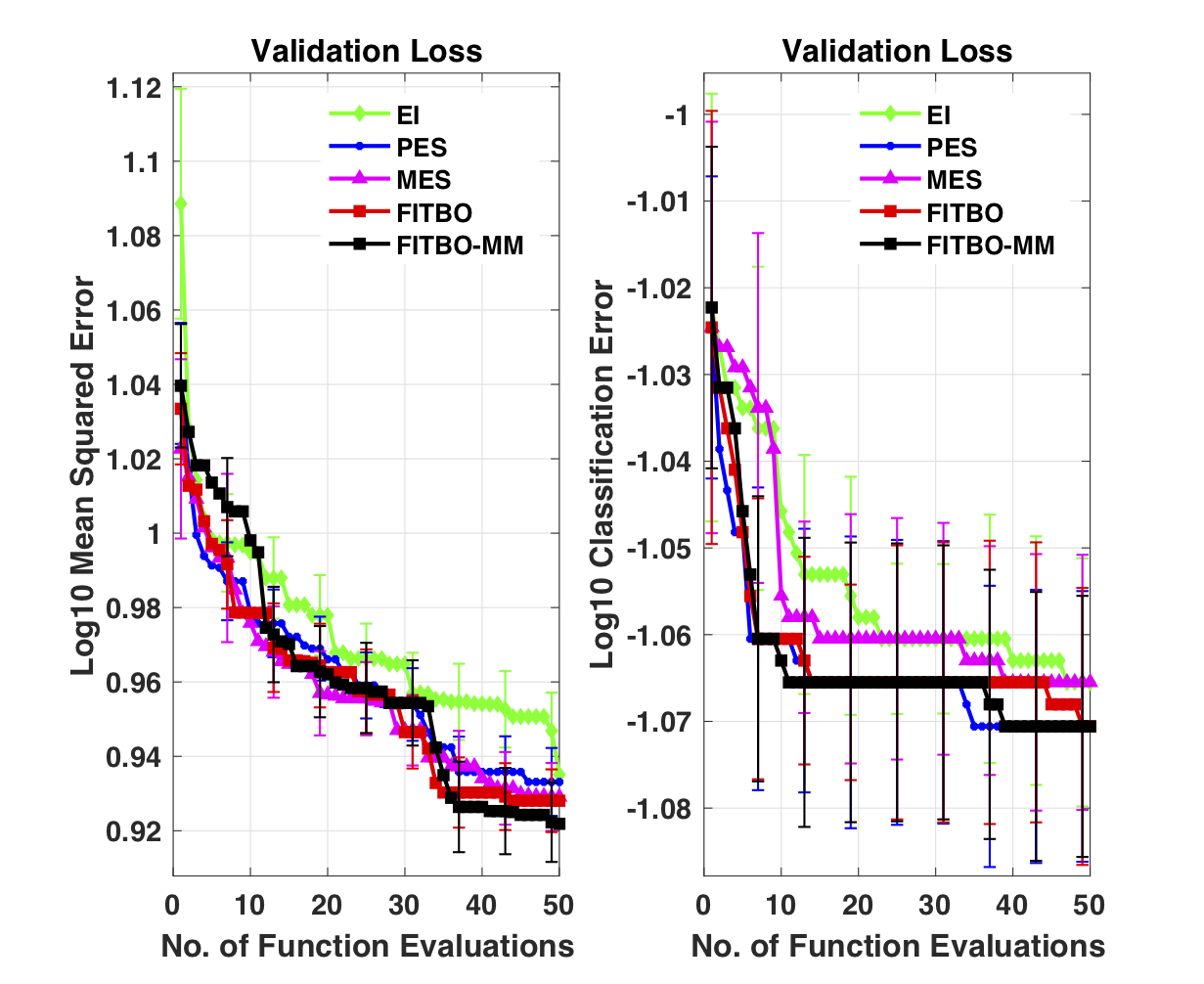}
        		   \caption{MNIST-SVM}
		   \label{MNISTSVM}
        \end{subfigure}
	\caption{Performance on tuning hyperparameters for (a) training a neural network on the Boston housing dataset and (b) training an SVM on the MNIST dataset.} 		   
	\label{Boston2D_MNIST2D}
\end{figure}

\begin{figure} [t] 
	 \centering
          \includegraphics[trim=0.4cm 0.1cm 0.8cm  0.1cm, clip, width=1.0\linewidth]{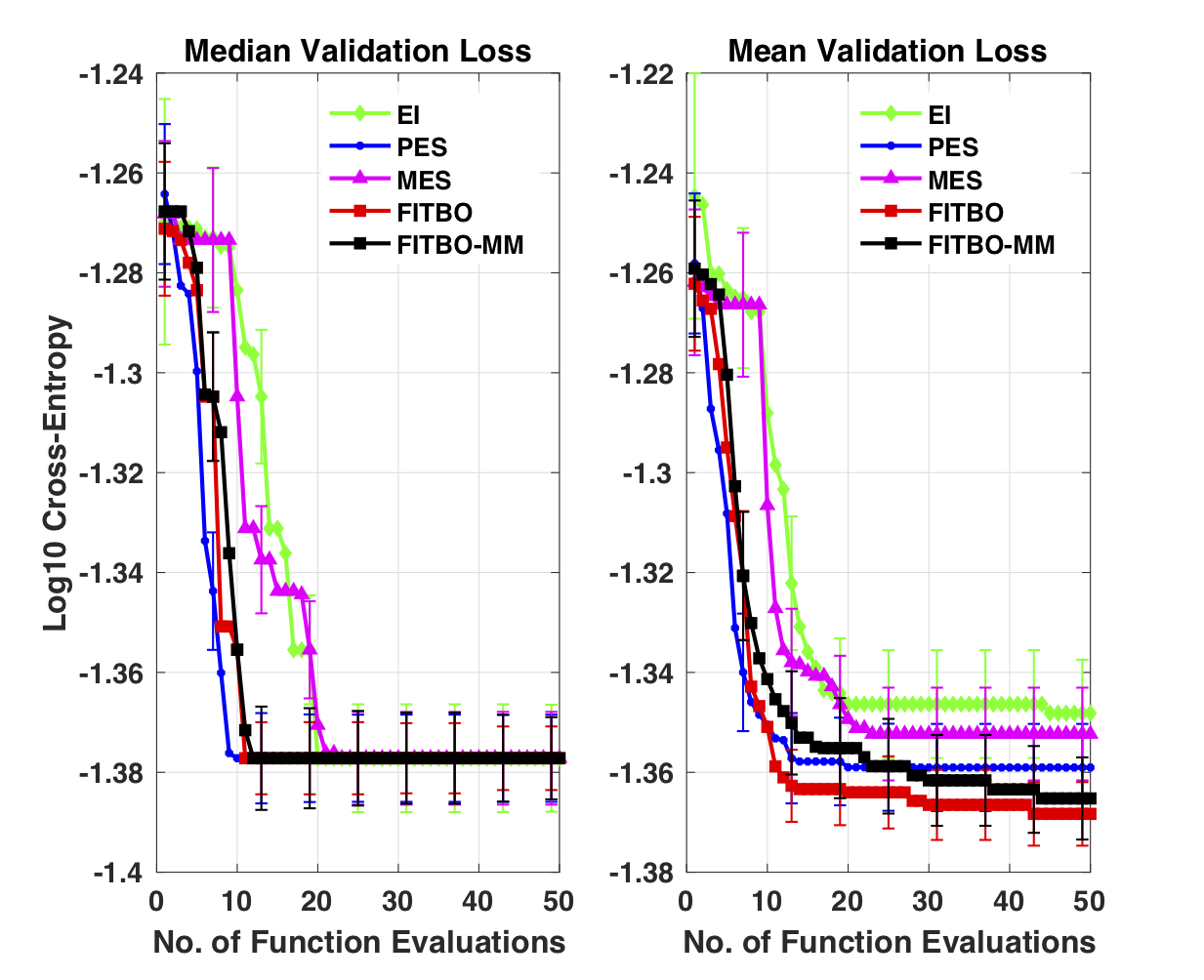}
	\caption{Performance on tuning hyperparameters for training a classification neural network on the breast cancer dataset.} 		   
	\label{Cancer4D}
\end{figure}

\smallbreak
We initialise all Bayesian optimisation algorithms with 3 random observation data and set the observation noise to $\sigma_n^2 = 10^{-3}$. All experiments are repeated 40 times. In each case, the ground truth is unknown but our aim is to minimise the validation loss. Thus, the corresponding loss functions are used to compare the performance of various Bayesian optimisation algorithms. 
\\\\
Figure \ref{Boston2D_MNIST2D} shows the median of the best validation losses achieved by all Bayesian optimisation algorithms after $n$ iterations for the Boston and MNIST-SVM problems. Our FITBO and FITBO-MM perform competitively well compared to their information-theoretic counterparts and all information-theoretic methods outperform EI in these real-world applications. 
\\\\
As for the Cancer problem (Figure \ref{Cancer4D}), FITBO and FITBO-MM converge to the stable median value of the validation loss at a much faster speed than MES and EI and are almost on par with PES. By examining the mean validation loss shown in the right plot of Figure \ref{Cancer4D}, it is evident that both FITBO and FITBO-MM demonstrate better performance than all other methods on average with FITBO gaining a slight advantage over FITBO-MM. Moreover, the comparable performance of FITBO and FITBO-MM in all three real-world tasks re-affirmed the robustness of our approach to entropy approximation as our moment matching technique, while improving the speed of the algorithm, does not really compromise the performance. 

\section{Conclusion}
We have proposed a novel information-theoretic approach for Bayesian optimisation, FITBO. With the creative use of the parabolic transformation and the hyperparameter $\eta$, FITBO enjoys the merits of less sampling effort, more flexible kernel choices and much simpler implementation in comparison with other information-based methods like PES and MES. As a result, its computational speed outperforms current information-based methods by a large margin and even exceeds EI to be on par with PI and GP-UCB. While requiring much lower runtime, it still manages to achieve satisfactory optimisation performance which is as good as or even better than PES and MES in a variety of tasks. Therefore, FITBO approach offers a very efficient and competitive alternative to existing Bayesian optimisation approaches.  

\section*{Acknowledgements}
We wish to thank Roman Garnett and Tom Gunter for the insightful discussions and Zi Wang for sharing the Matlab implementation of EI, PI, GP-UCB, MES and PES. We would also like to thank Favour Mandanji Nyikosa, Logan Graham, Arno Blaas and Olga Isupova for their helpful comments about improving the paper. 


%
\bibliographystyle{abbrvnat}
\bibliography {library}

\begin{thebibliography}{21}
\providecommand{\natexlab}[1]{#1}
\providecommand{\url}[1]{\texttt{#1}}
\expandafter\ifx\csname urlstyle\endcsname\relax
  \providecommand{\doi}[1]{doi: #1}\else
  \providecommand{\doi}{doi: \begingroup \urlstyle{rm}\Url}\fi

\bibitem[Bache and Lichman(2013)]{bache2013uci}
K.~Bache and M.~Lichman.
\newblock {UCI} machine learning repository.
\newblock 2013.

\bibitem[Bochner(1959)]{bochner1959lectures}
S.~Bochner.
\newblock \emph{Lectures on Fourier Integrals: With an Author's Suppl. on
  Monotonic Functions, Stieltjes Integrals and Harmonic Analysis. Transl. from
  the Orig. by Morris Tennenbaum and Harry Pollard}.
\newblock University Press, 1959.

\bibitem[Brochu et~al.(2010)Brochu, Cora, and De~Freitas]{brochu2010tutorial}
E.~Brochu, V.~M. Cora, and N.~De~Freitas.
\newblock A tutorial on {B}ayesian optimization of expensive cost functions,
  with application to active user modeling and hierarchical reinforcement
  learning.
\newblock \emph{arXiv preprint arXiv:1012.2599}, 2010.

\bibitem[Gunter et~al.(2014)Gunter, Osborne, Garnett, Hennig, and
  Roberts]{gunter2014sampling}
T.~Gunter, M.~A. Osborne, R.~Garnett, P.~Hennig, and S.~J. Roberts.
\newblock Sampling for inference in probabilisktic models with fast {B}ayesian
  quadrature.
\newblock In \emph{Advances in neural information processing systems}, pages
  2789--2797, 2014.

\bibitem[Hennig and Schuler(2012)]{hennig2012entropy}
P.~Hennig and C.~J. Schuler.
\newblock Entropy search for information-efficient global optimization.
\newblock \emph{Journal of Machine Learning Research}, 13\penalty0
  (Jun):\penalty0 1809--1837, 2012.

\bibitem[Hern{\'a}ndez-Lobato et~al.(2014)Hern{\'a}ndez-Lobato, Hoffman, and
  Ghahramani]{hernandez2014predictive}
J.~M. Hern{\'a}ndez-Lobato, M.~W. Hoffman, and Z.~Ghahramani.
\newblock Predictive entropy search for efficient global optimization of
  black-box functions.
\newblock In \emph{Advances in neural information processing systems}, pages
  918--926, 2014.

\bibitem[Hoffman and Ghahramani(2015)]{hoffman:2015}
M.~W. Hoffman and Z.~Ghahramani.
\newblock Output-space predictive entropy search for flexible global
  optimization.
\newblock In \emph{the NIPS workshop on {B}ayesian optimization}, 2015.

\bibitem[Huber et~al.(2008)Huber, Bailey, Durrant-Whyte, and
  Hanebeck]{huber2008entropy}
M.~F. Huber, T.~Bailey, H.~Durrant-Whyte, and U.~D. Hanebeck.
\newblock On entropy approximation for {G}aussian mixture random vectors.
\newblock In \emph{Multisensor Fusion and Integration for Intelligent Systems,
  2008. MFI 2008. IEEE International Conference on}, pages 181--188. IEEE,
  2008.

\bibitem[Jones et~al.(1998)Jones, Schonlau, and Welch]{jones1998efficient}
D.~R. Jones, M.~Schonlau, and W.~J. Welch.
\newblock Efficient global optimization of expensive black-box functions.
\newblock \emph{Journal of Global optimization}, 13\penalty0 (4):\penalty0
  455--492, 1998.

\bibitem[Kandasamy et~al.(2015)Kandasamy, Schneider, and
  P{\'o}czos]{kandasamy2015high}
K.~Kandasamy, J.~Schneider, and B.~P{\'o}czos.
\newblock High dimensional {B}ayesian optimisation and bandits via additive
  models.
\newblock In \emph{International Conference on Machine Learning}, pages
  295--304, 2015.

\bibitem[Kushner(1964)]{kushner1964new}
H.~J. Kushner.
\newblock A new method of locating the maximum point of an arbitrary multipeak
  curve in the presence of noise.
\newblock \emph{Journal of Basic Engineering}, 86\penalty0 (1):\penalty0
  97--106, 1964.

\bibitem[LeCun et~al.(1998)LeCun, Bottou, Bengio, and
  Haffner]{lecun1998gradient}
Y.~LeCun, L.~Bottou, Y.~Bengio, and P.~Haffner.
\newblock Gradient-based learning applied to document recognition.
\newblock \emph{Proceedings of the IEEE}, 86\penalty0 (11):\penalty0
  2278--2324, 1998.

\bibitem[Mo{\v{c}}kus et~al.(1978)Mo{\v{c}}kus, Tiesis, and
  {\v{Z}}ilinskas]{movckus1978toward}
J.~Mo{\v{c}}kus, V.~Tiesis, and A.~{\v{Z}}ilinskas.
\newblock Toward global optimization, volume 2, chapter the application of
  {B}ayesian methods for seeking the extremum, 1978.

\bibitem[Murray et~al.(2010)Murray, Prescott~Adams, and
  MacKay]{murray2010elliptical}
I.~Murray, R.~Prescott~Adams, and D.~J. MacKay.
\newblock Elliptical slice sampling.
\newblock 2010.

\bibitem[Rasmussen and Williams(2006)]{rasmussen2006gaussian}
C.~E. Rasmussen and C.~K. Williams.
\newblock \emph{Gaussian processes for machine learning}, volume~1.
\newblock MIT press Cambridge, 2006.

\bibitem[Requeima(2016)]{requeimaintegrated}
J.~R. Requeima.
\newblock Integrated predictive entropy search for {B}ayesian optimization.
\newblock 2016.

\bibitem[Shahriari et~al.(2016)Shahriari, Swersky, Wang, Adams, and
  de~Freitas]{shahriari2016taking}
B.~Shahriari, K.~Swersky, Z.~Wang, R.~P. Adams, and N.~de~Freitas.
\newblock Taking the human out of the loop: A review of {B}ayesian
  optimization.
\newblock \emph{Proceedings of the IEEE}, 104\penalty0 (1):\penalty0 148--175,
  2016.

\bibitem[Snoek et~al.(2012)Snoek, Larochelle, and Adams]{snoek2012practical}
J.~Snoek, H.~Larochelle, and R.~P. Adams.
\newblock Practical {B}ayesian optimization of machine learning algorithms.
\newblock In \emph{Advances in neural information processing systems}, pages
  2951--2959, 2012.

\bibitem[Srinivas et~al.(2009)Srinivas, Krause, Kakade, and
  Seeger]{srinivas2009gaussian}
N.~Srinivas, A.~Krause, S.~M. Kakade, and M.~Seeger.
\newblock Gaussian process optimization in the bandit setting: No regret and
  experimental design.
\newblock \emph{arXiv preprint arXiv:0912.3995}, 2009.

\bibitem[Villemonteix et~al.(2009)Villemonteix, Vazquez, and
  Walter]{villemonteix_informational_2009}
J.~Villemonteix, E.~Vazquez, and E.~Walter.
\newblock An informational approach to the global optimization of
  expensive-to-evaluate functions.
\newblock \emph{Journal of Global Optimization}, 44\penalty0 (4):\penalty0
  509--534, 2009.
\newblock URL \url{http://www.springerlink.com/index/T670U067V47922VK.pdf}.

\bibitem[Wang and Jegelka(2017)]{wang2017max}
Z.~Wang and S.~Jegelka.
\newblock Max-value entropy search for efficient {B}ayesian optimization.
\newblock \emph{arXiv preprint arXiv:1703.01968}, 2017.

\end{thebibliography}

\end{document}